\documentclass[conference]{IEEEtran}
\IEEEoverridecommandlockouts
\usepackage{cite}
\usepackage{amsmath,amssymb,amsfonts}
\usepackage{algorithmic}
\usepackage{graphicx}
\usepackage{textcomp}
\usepackage{xcolor}
\usepackage{tcolorbox}
\usepackage{booktabs} 
\usepackage{multirow}

\def\BibTeX{{\rm B\kern-.05em{\sc i\kern-.025em b}\kern-.08em
    T\kern-.1667em\lower.7ex\hbox{E}\kern-.125emX}}

\begin{document}

\title{Exploring the Implicit Semantic Ability of Multimodal Large Language Models: A Pilot Study on Entity Set Expansion
}

\author{\IEEEauthorblockN{1\textsuperscript{st} Hebin Wang$^{\dagger}$}
\IEEEauthorblockA{\textit{Shenzhen International Graduate School} \\
\textit{Tsinghua University}}
\and
\IEEEauthorblockN{2\textsuperscript{nd} Yangning Li$^{\dagger}$}
\IEEEauthorblockA{\textit{Shenzhen International Graduate School} \\
\textit{Tsinghua University, Peng Cheng Laboratory}}
\and
\IEEEauthorblockN{3\textsuperscript{rd} Yinghui Li}
\IEEEauthorblockA{\textit{Shenzhen International Graduate School} \\
\textit{Tsinghua University}}
\and
\IEEEauthorblockN{4\textsuperscript{th} Hai-Tao Zheng}
\IEEEauthorblockA{\textit{Shenzhen International Graduate School} \\
\textit{Tsinghua University, Peng Cheng Laboratory}}
\and
\IEEEauthorblockN{5\textsuperscript{th} Wenhao Jiang$^{*}$}
\IEEEauthorblockA{\textit{Guangdong Laboratory of Artificial Intelligence and Digital Economy (SZ)}}
\and
\IEEEauthorblockN{6\textsuperscript{th} Hong-Gee Kim}
\IEEEauthorblockA{\textit{Seoul National University}}
\thanks{$^{\dagger}$ Equally contribution
}
\thanks{* Corresponding author. (E-mail: cswhjiang@gmail.com) This research is supported by National Natural Science Foundation of China(Grant No.62276154), Research Center for ComputerNetwork (Shenzhen) Ministry of Education, the Natural Science Foundation of Guangdong Province(Grant No.2023A1515012914 and 440300241033100801770), Basic Research Fund of Shenzhen City (Grant No.JCYJ20210324120012033, JCYJ20240813112009013 and GJHZ20240218113603006), the Major Key Projectof PCL for Experiments and Applications (PCL2023A09).
}
}
%

\maketitle

\begin{abstract}
The rapid development of multimodal large language models (MLLMs) has brought significant improvements to a wide range of tasks in real-world applications. However, LLMs still exhibit certain limitations in extracting implicit semantic information. In this paper, we applies MLLMs to the Multi-modal Entity Set Expansion (MESE) task, which aims to expand a handful of seed entities with new entities belonging to the same semantic class, and multi-modal information is provided with each entity. We explore the capabilities of MLLMs to understand implicit semantic information at the entity-level granularity through the MESE task, introducing a listwise ranking method LUSAR that maps local scores to global rankings. Our LUSAR demonstrates significant improvements in MLLM's performance on the MESE task, marking the first use of generative MLLM for ESE tasks and extending the applicability of listwise ranking.

\end{abstract}

\begin{IEEEkeywords}
Entity Set Expansion, Multimodal Large Language Models, Implicit Semantic Reasoning
\end{IEEEkeywords}

\section{Introduction}

Large language models (LLMs) have demonstrated outstanding performance across a wide range of NLP tasks~\cite{li2024correct, DBLP:conf/aaai/YangMZWL020, li2022past, ye2023cleme, li2022learning, ma2022linguistic} since their inception~\cite{yang2024execrepobench, yang2024evaluating,  li2024llms, li2024rethinking, du2024llms, huang2023lateval, li2023effectiveness,li2023active}. 
The Multi-modal Language Models (MLLMs) merge the advanced reasoning capabilities of traditional LLMs with the processing of image, video, and audio data, garnering widespread attention for its ability to understand and generate visual language~\cite{liu2022we, li2023towards, li2024ecomgpt,li2024mesed }. 

Despite the remarkable capabilities of MLLMs, they exhibit notable limitations when it comes to implicit semantic information extraction and reasoning. One of the challenges for MLLMs lies in their reliance on explicit contextual information. Without well-defined prompts or rich context, MLLMs struggle to infer hidden semantic relationships. For MLLMs, the fusion of different modalities can lead to an overemphasis on primary features, such as dominant visual cues in image-text tasks, while neglecting subtler yet important contextual signals. This issue limits their ability to infer complex relationships between entities across modalities. 


In our work, we explore the implicit semantic information extraction capabilities of MLLMs by applying the Multimodal Entity Set Expansion (MESE) task as probing task, which is a representative task for evaluating their ability to infer hidden semantics. The MESE task ~\cite{huang2023retrieval,li2023automatic,li2024mesed} seeks to expand the given set of seed entities by identifying additional entities that belong to the same semantic class, utilizing a predefined candidate entity vocabulary and corpus. For instance, given the multimodal seed entities {\textit{New York}, \textit{Chicago}, \textit{Los Angeles}}, the goal of MESE is to retrieve other entities that implicitly share the semantic category \texttt{US Cities}, such as \textit{NYC}, and \textit{Boston}. 

The challenge of the MESE task lies in the implicit semantic induction of entity set, which aligns with the implicit semantic information extraction challenges of MLLMs. Therefore the performance on the MESE task can serves as an indicator of the model's effectiveness in uncovering semantic information at the entity level.
Taking the challenge of negative entities with fine-grained semantic differences in the MESE task as an example, without an explicit prompt specifying the exact semantic class of the three seed entities, MLLM may encounter difficulties in inferring the hidden common features from a very small number of instances, in which they tend to generalize broad and imprecise semantic classes, rather than extracting more fine-grained features. For instance, when provided with a few plant entities (such as \textit{oak} and \textit{maple}) and tasked with inferring that they belong to the category \texttt{Deciduous Trees}, the model may fail to deduce this hidden class, while mistakenly thought to be a coarser-grained semantic class \texttt{Trees}. 


To address this issue and facilitate implicit reasoning in MLLMs, we proposed a \textbf{L}istwise m\textbf{U}ltimodal \textbf{S}ampling \textbf{A}nd \textbf{R}anking method (\textbf{LUSAR}). In MESE task, the listwise LUSAR amplifies the differences between coarse- and fine-grained candidate entities by comparing their relative semantic similarity to the seed entities' class. By leveraging the comparison of semantic class similarities, this approach identifies candidate entities that are closer to fine-grained positive entities, addressing the challenge where large models struggle to directly infer and generalize fine-grained semantic features from implicit information. This listwise approach overcomes the limitations of MLLMs in insufficiently extracting hidden semantic information from limited entity instances.

While traditional listwise approaches are applied on document retrieval\cite{xia2008listwise,cao2007learning}, they focus on probabilistic method to ranking problems. 
We established a generalizable LLM listwise application paradigm, where uniform sampling is employed to use the local ranking score of elements to reflect their overall ranking positions. 
Our experiments indicate that application of the listwise approach to the MESE task has brought excellent performance improvement to MLLM, making a big enhancement on MLLM's implicit semantic information inference. 

In summary, our work makes the following key contributions:
\begin{enumerate}
    \item We are the first to apply MLLM to the MESE task, exploring their ability to understand implicit semantics and developing their capabilities in MESE.
    \item We propose a scoring and ranking mechanism centered around the listwise approach termed LUSAR for the MESE task, and innovatively integrate the listwise mechanism into MLLMs. This not only enhances the semantic understanding of large models but also extends the application of the listwise approach within large models, offering broader prospects in tasks such as recommendation systems.
    \item Extensive experiments demonstrate the effectiveness of our proposed method. With our improvements, MLLMs show promising performance gains in the MESE task.
\end{enumerate}

\section{Methodology}

\begin{figure}
\centering
\scalebox{0.6}{
\centerline{\includegraphics[width=0.8\textwidth]{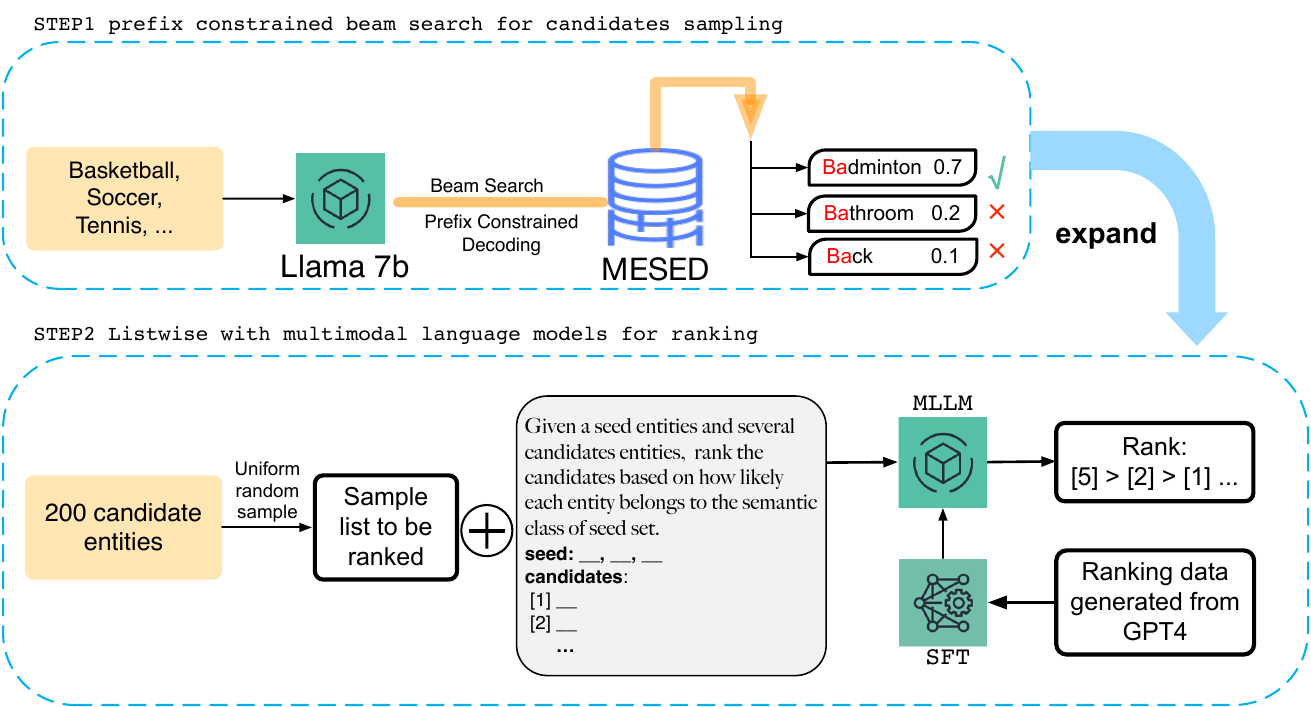}}}
\caption{Overall framework of our LUSAR.}
\vspace{-0.5cm}
\label{framework}
\end{figure}

We propose a novel method LUSAR, which is divided into two stages. As illustrated in Figure \ref{framework}, in the first stage, a prefix tree constraint is used to restrict the large model to generate a large number of candidate entities within a specified entity dataset. In the second stage, we designed a listwise approach. This approach scientifically performs multiple sampling and ranking of each candidate entity to obtain a ranking score for each entity. The sorted scores are then used as the final order for evaluation.
\subsection{Candidates Entities Sample}

We employ a generative framework with constrained
decoding strategy to the large language model, generating a certain number of candidate entities for the initial step of entity filtering.
Instead of relying on the intermediate semantic features of entities as traditional models do, we adopt a generative method to accomplish entity set expansion. To ensure that the LLM generates only valid entities within the entity vocabulary, we build a prefix tree based on the entity vocabulary and employ prefix-constrained Beam Search for decoding\cite{sutskever2014sequence,huang2024retrieval}. 
During the generation, at each timestep the model extend a patial entity in the beam  words with every possible word in the vocabulary. Beam search will only maintain the top \textit{n} most likely tokens based on the model’s log probability, where \textit{n} represents for the number of beams. Once the <EOS> token is appended to a entity, it is removed from the beam and added to the set of completed candidate entities set.
\subsection{Listwise Ranking prompting}
\subsubsection{prompting design}
\par

We utilize natural language prompt templates to transform the ranking problem into a generation problem. The large language model will process the generation task and only output the sequence of entities ranked by their relevance to the semantic class represented by the seed entity. 

We use $R_m=r(s,E_m)$ to denote the the output ranked sequence, where $s$ represents the seed entity set \{A,B,C\} used to query, $E_m$ is the canditate entities $e_1,e_2,...e_n$. $n$ is  is typically set to 5, to balance the increasing entity relevance with the number of entities involved in the ranking, and the model's ability to process image information. $R_m$ is the ranked order of canditate entities, $\{R_m\}=\{E_m\}$.
{~}

\subsubsection{List Comparisons and Scoring}

While pairwise approaches require to enumerate all pairs and perform a global aggregation to give each item a score, it is impractical for the listwise approach to generate and rank every possible combination of all candidate entities using LLM.

We developed an algorithm that samples an appropriate number of lists from the entire set of candidate entities, ensuring that each entity appears with balanced frequency. We set the sample size $n$ to 5, which means LLM ranks five candidate entities in each query. And the occurrences each entity appears $o$ is set to 10. In the whole candidate entities set, each entity appears in ten sample lists.









\begin{table}[]
\centering
\scalebox{0.8}{
\begin{tabular}{|p{10cm}|}
\hline
\textbf{Instruction:}\\
Given an input seed entity set. The set defines a particular semantic class based on the common attributes of its members. Then input 5-6 candidate entities. Each entity is combined with a corresponding image. Your task is to generate a ranking of the provided candidate entities, based on their relevance to the seed entity set(the probability that each entity belongs to the same semantic class of the seed set). 

Output the five candidate entities with ranking order, from the most relevant to least relevant, without regard to the initial order of the candidates.

\medskip

\noindent
\textbf{Input:}\\
\texttt{[Seed entity set with images]:}\\
\hspace*{2em}\{A\}\{image A\}\\
\hspace*{2em}\{B\}\{image B\}\\
\hspace*{2em}\{C\}\{image C\}\\
\texttt{[Candidate entities with images]:}\\
\hspace*{2em}\{D\}\{image D\}\\
\hspace*{2em}\{F\}\{image F\}\\
\hspace*{2em}\ldots

\noindent
\textbf{Response:}\\
\hspace*{2em} I \textgreater  F \textgreater  D \ldots
\\ \hline
\end{tabular}}
\title{Listwise Prompt}
\vspace{-0.5cm}
\label{prompt}
\end{table}

The random grouping mechanism, combined with the repeated inclusion of each entity, allows for a comprehensive consideration of the relativity between entities of varying quality within the context of randomized ranking lists.
In probability theory, with appropriate sample size, sample set can represent the statistical properties of the population, and the performance within the sample can be regarded as indication of the overall population level. Consequently, the performance score of a single entity across these lists can reflect its relative position within the entire set of candidate entities.

Our listwise approach designed for LLM transforms the traditional method of utilizing ranking to reflect probabilities into a scoring mechanism on LLM. After random sampling, it is ensured that each ranking list contains $n$ candidate entities, with each candidate entity appearing in $o$ lists. After all the comparisons completed, we conduct a global aggregation to assign a score $score_e$ for each entity. Specifically, we have:

\begin{gather}
R_m=r(s,E_m) \\
R_m=[r_{m1},r_{m2},...r_{mn}] \\
score_{e_k}=\sum_{m=1}^M \sum_{i=1}^n i\cdot\mathbb{I}_{e_k=r_{mi}}
\end{gather}

where $M$ denotes the the total number of sample lists, $M=o*NUM_{candidates}/n$. $n$ is the length of single sample list. $R_m$ is the ordered list resulting from the ranking of the candidate subset $E_m$, where $r_{mi}$ represents the entity ranked in the i-th position within $E_m$, with $1 \leq i \leq n$.

\begin{table*}[]
\centering
\caption{Main experiment results. Text-based, vision-based, Generative MLLM and Listwise MLLM expansion methods are evaluated. }
\scalebox{0.8}{
\begin{tabular}{clccccccccc}
\toprule
                              \multirow{2}{*}{\textbf{Model Type}}     & \multirow{2}{*}{\textbf{Model}}                                 & \multicolumn{4}{c}{\textbf{MAP}}                           & \multicolumn{4}{c}{\textbf{P}}                             & \multirow{2}{*}{\textbf{Avg}} \\ \cmidrule{3-10}
                                   &                                  & \textbf{@10} & \textbf{@20} & \textbf{@50} & \textbf{@100} & \textbf{@10} & \textbf{@20} & \textbf{@50} & \textbf{@100} &                               \\ \midrule
\multirow{6}{*}{\begin{tabular}[c]{@{}l@{}}Text or Vision\\-based model\\tailored for ESE\end{tabular}}        & SetExpan                & 26.10 & 20.98 & 15.83 & 13.91 & 34.25 & 29.58 & 24.25 & 22.96 & 23.48                \\ \cmidrule{2-11} 
                          
                          & CGExpan                 & 38.89 & 32.51 & 24.69 & 21.06 & 45.85 & 39.85 & 33.19 & 32.80 & 33.61                \\ \cmidrule{2-11} 
                          
                          & ProbExpan               & 65.47 & 57.50 & 43.96 & 40.73 & 71.30 & 64.35 & 55.73 & 51.99 & 56.38                \\ \cmidrule{2-11} 
      & CLIP                    & 76.41 & 65.75 & 49.58 & 40.08 & 79.20 & 69.53 & 53.10 & 43.66 & 59.66                \\ \cmidrule{2-11} 
                          & ALBEF                   & 83.55 & 75.46 & 63.02 & 54.47 & 86.60 & 79.15 & 68.03 & 61.12 & 71.43                \\ \midrule
\multirow{2}{*}{\begin{tabular}[c]{@{}l@{}}Generative MLLM\end{tabular}}      & Qwen-VL-Chat                    & 52.07 & 47.65 & 42.57 & 40.60 & 65.53 & 62.03 & 58.69 & 63.14 & 53.78                \\ \cmidrule{2-11} 
                          & Deepseek-VL-7b                   & 69.88 & 66.30 & 63.74 & 63.88 & 86.60 & 79.15 & 68.03 & 61.12 & 69.59                \\ \midrule 
           \multirow{3}{*}{\begin{tabular}[c]{@{}l@{}}Our LUSAR\end{tabular}}                & \textbf{Qwen-VL-Chat }       & 68.45 & 64.60 & 61.88 & 61.95 & 78.56 & 75.05 & 75.51 & 82.45 & 71.06                \\ \cmidrule{2-11} 
                          & \textbf{Qwen2-VL-7b  }       & 82.15 & 77.62 & 75.24 & 74.77 & 88.31 & 85.03 & 85.91 & 88.33 & \textbf{83.17}                \\ \cmidrule{2-11} 
                          & \textbf{Deepseek-VL-7b}       & 76.58 & 71.79 & 68.13 & 67.44 & 84.21 & 80.79 & 80.91 & 84.36 & 76.90 \\  \bottomrule

\end{tabular}}
\vspace{-0.5cm}
\label{tab:main}
\end{table*}

\subsubsection{Listwise Fine-tune}
In the context of the listwise approaches, where the goal is to rank the likelihood that candidate entities belong to the semantic class of the seed entity set-predict $R_m$ given $(s,E_m)$. We observed that LLM has difficulty in performing ranking task, We employ a supervised fine-tuning strategy utilizing the Low-Rank Adaptation (LoRA) technique, as introduced by \cite{hu2021lora}.

We use GPT-4 to generate a batch of data for SFT. Each data entry includes a seed entity set belonging to the same semantic class, several candidate entities, and the ground truth ranking position of each within the candidate set. We intentionally included a varying number of distractor candidiate entities to each data entry that do not belong to the target semantic class, therefore to train the model’s discriminative ability and enhance its distinction between positive and negative examples. For multimodal information, we crawled representative images of each entity from Google and incorporated them into the input for fine-tuning. The overall data size after filtering and processing is approximately 4000 entries. To prevent the degradation of MLLMs and preserve their original question-answering capabilities, we integrated a certain proportion of 
LLaVA-Instruct-150K \cite{liu2024improved} to fine-tune.
The fine-tuning process using GPT4 data rectified issues in the LLM, such as repeated content, non-compliant formatting, and irrelevant responses, thereby improving the model's performance on the listwise ranking task.


\section{Experiments}
\label{methodology}

\subsection{Dataset}

For our experiments, we use the MESED\cite{li2024mesed} dataset as the benchmark, which is a multi-modal ESE dataset with meticulous manual calibration. The MESED comprises 70 fine-grained semantic classes, with each fine-grained class contains 5 queries with three seed entities and 5 queries with five seed entities.

\subsection{Baseline Models}
We compare two categories of baseline model, the first is the (multimodal) LM-based model tailored for ESE, including \textbf{SetExpan}~\cite{shen2017setexpan}, \textbf{CGExpan}~\cite{zhang2020empower}, \textbf{ProbExpan}~\cite{li2022contrastive}, \textbf{CLIP}~\cite{radford2021learning}, and \textbf{ALBEF}~\cite{li2021align}. Of the above models, the former three are the are based on pre-trained language model BERT. CLIP and ALBEF are multimodal ESE models with text and images as inputs. The other category of models are generative MLLMs with instruction-following capabilities, which is \textbf{Qwen-VL-Chat}\cite{bai2023qwen}, \textbf{Deepseek-7b-VL-Chat}\cite{lu2024deepseek}, \textbf{Qwen2-VL-7b-Instruct}\cite{yang2024qwen2}. To accommodate the number of images associated with multiple seed entities and candidate entities in the MESE task, only multimodal large models that support multiple images were selected.

\noindent\textbf{Evaluation Metrics} The objective of MESE is to expand the ranked entity list based on their similarity to given seed entities in descending order. Two widely used evaluation metrics, MAP@$K$ and P@$K$, are employed, also utilized in previous research \cite{li2024mesed,zhang2020empower}. The MAP@$K$ metric is computed as follows:
\begin{equation}
    \mbox{MAP@}K=
    \frac{1}{|Q|}\sum_{q \in Q}
    \mbox{AP}_K(R_q,G_q)
\end{equation}
Here, $Q$ is the collection for each query $q$. $\mbox{AP}_K(R_q,G_q)$ denotes the average precision at position $K$ with the ranked list $R_q$ and ground-truth list $G_q$. P@$K$ is the precision of the top-$K$ entities. In the experiment, queries with $\|$Seed$\|$=3 and 5 are evaluated separately.
\subsection{Backbone MLLM for LUSAR}
Our baseline use Qwen-VL-Chat model\cite{bai2023qwen}. To contrast with our listwise approach, we use single MLLM to complete the task of scoring each entity, eliminating the comparison between multiple candidate entities. The relative nature of the listwise method is removed, and an absolute scoring approach is adopted, based on the similarity of each individual entity to the seed entities.

\subsection{Main Experiments}
The results of the main experiment are presented in Table \ref{tab:main}, from which we observe that: (1) The multi-modal methods outperform the mono-modal (text or vision based model) methods in general. The multimodal entity set expansion task incorporates image information in addition to text. The introduction of images provides an additional dimension for the model to uncover the latent semantic information of entities. By utilizing representative images, the model can effectively alleviate challenges associated with synonyms, polysemy, and hard negative entities.

(2) The capabilities of multimodal large models in entity set expansion tasks are limited, with baseline methods performing worse than traditional models. By introducing the listwise approach, we unlocked the potential of large models in the ESE task, resulting in significant performance improvements. The introduction of the listwise mechanism enhances the relative comparison between different candidate entities. The method of scoring entities through multiple sampling reflects their relative ranking within the global candidate set. The listwise approach efficiently addresses the challenge of global ranking by utilizing comparisons of shorter sequences in multiple iterations, achieving high performance.

\subsection{Ablation Study}

\begin{figure}[htbp]
  \centering
  \begin{minipage}[b]{0.5\linewidth} 
    \centering
    \includegraphics[width=\linewidth]{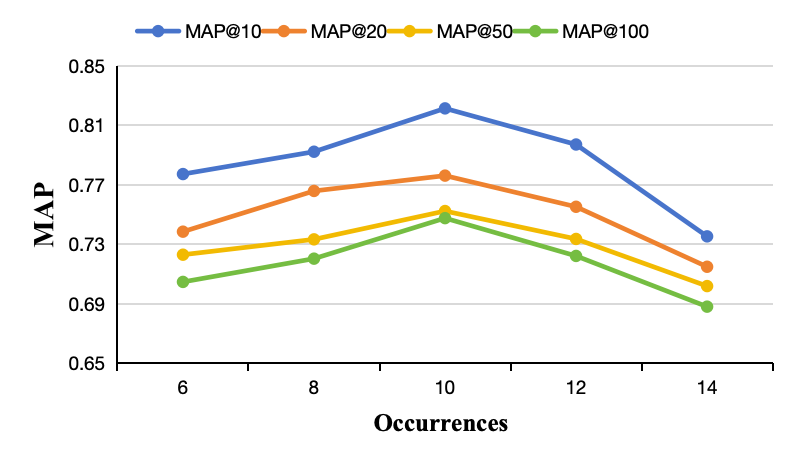} 
    \label{fig:image1}
  \end{minipage}
  \begin{minipage}[b]{0.4\linewidth} 
    \centering
    \includegraphics[width=\linewidth]{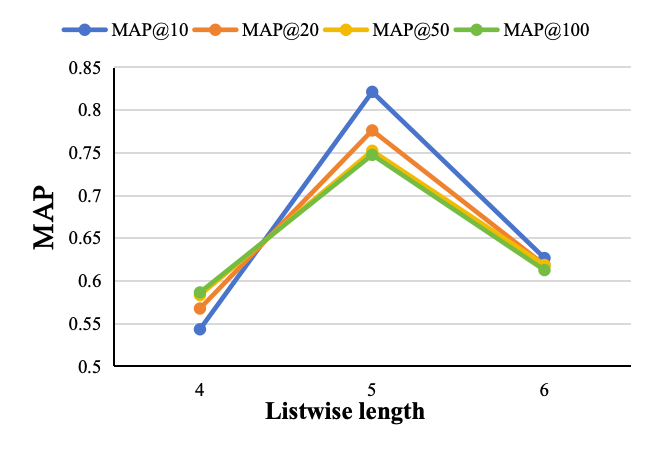} 
    \label{fig:image2}
  \end{minipage}
  \vspace{-0.5cm}
  \caption{Analysis of varying entity occurrences and sample sizes.}
  \vspace{-0.25cm}
\end{figure}
We conducted experiments on the frequency of different candidate entities and the length of individual ranking sequences in the Listwise method. Our Listwise mechanism utilizes uniform random sampling, where the scores obtained from sample rankings represent the overall ranking. Therefore, the length of the listwise sequences and the frequency of entity appearances are crucial to ensure that the sample ranks and scores reflects the overall statistical pattern in probabilistic theory.

Our ablation study shows that the experimental setup with occurrences $o=10$ and a listwise sample size $n=5$ yields the best performance. Under these parameter settings, the scores obtained by the candidate entities in the sample sequence ranking most accurately reflect their ground truth positions in the global sequence ranking.

\section{Conclusion}
To explore the implicit semantic ability of the multimodal large language models, we conduct a pilot study on MESE task. We propose LUSAR - a novel listwise paradigm on LLM to MESE task. . Extensive experiments demonstrate the significance of our method in enhancing MLLMs' ability to infer implicit semantic information, resulting in substantial performance improvements on MESE task metrics. Moreover, our method is the first to apply LLMs to the ESE task. Our LUSAR framework innovatively introduces a new paradigm for utilizing listwise ranking in large models, with broader applicability across various domains.

\bibliographystyle{IEEEtran}
\bibliography{reference}
\vspace{12pt}

\end{document}